\documentclass[10pt,letterpaper]{article}
\usepackage[top=0.85in,left=2.75in,footskip=0.75in]{geometry}

\usepackage{changepage}

\usepackage[utf8]{inputenc}

\usepackage{textcomp,marvosym}

\usepackage{fixltx2e}

\usepackage{amsmath,amssymb}

\usepackage{cite}

\usepackage{nameref,hyperref}

\usepackage[right]{lineno}

\usepackage{microtype}
\DisableLigatures[f]{encoding = *, family = * }

\usepackage{rotating}

\usepackage{CJKutf8}

\raggedright
\usepackage[aboveskip=1pt,labelfont=bf,labelsep=period,justification=raggedright,singlelinecheck=off]{caption}

\makeatletter
\renewcommand{\@biblabel}[1]{\quad#1.}
\makeatother

\date{}

\usepackage{lastpage,fancyhdr,graphicx}
\usepackage{epstopdf}
\fancyheadoffset[L]{2.25in}
\fancyfootoffset[L]{2.25in}
\begin{document}

\begin{CJK*}{UTF8}{gbsn}
\vspace*{0.35in}

\begin{flushleft}
{\Large
\textbf\newline{Optimizing the Learning Order of Chinese Characters Using a Novel Topological Sort Algorithm}
}
\newline
\\
James C. Loach\textsuperscript{*},
Jinzhao Wang\textsuperscript{}
\\
\bigskip
INPAC and Dept. of Physics, Shanghai Jiao Tong University, Shanghai 200240, China.
\bigskip

%
%





* james.loach@sjtu.edu.cn

\end{flushleft}
\section*{Abstract}
We present a novel algorithm for optimizing the order in which Chinese characters are learned, one that incorporates the benefits of learning them in order of usage frequency and in order of their hierarchal structural relationships. We show that our work outperforms previously published orders and algorithms. Our algorithm is applicable to any scheduling task where nodes have intrinsic differences in importance and must be visited in topological order.



\section*{Introduction}

One of the most fascinating aspects of the Chinese language presents one of the largest barriers to learning it, an irony not lost on generations of students. Chinese characters enchant the learner like little else, but the origins of that enchantment, their elegant, structured complexity and seemingly infinite variety, make learning them a formidable task \cite{ke2001, walker1989}. Mastery is a hard-won thing and is rarely achieved until an advanced stage of study \cite{light1975}.

Becoming functionally literate in Chinese requires memorization of several thousand distinct characters, and the effort involved has profound consequences for the learning process \cite{kane}. An early focus on learning characters can delay the acquisition of productive language skills, while learning them late can inhibit productive learning techniques, such as extensive reading, and also obscure the logic of the language. Either way, the consequences for students are a steep learning curve, high rates of attrition, and a certain preoccupation with methods for learning and remembering characters \cite{marton1}.

The task of learning thousands of distinct symbols is not, however, as difficult as it first appears. There are regularities in the structures of Chinese characters that relate them to their pronunciations and meanings, and also to one another. Around 90\% of characters are semantic-phonetic compounds, in which one part of the character indicates the meaning and the other part the pronunciation. These cues are not always obvious, as meanings and pronunciations have evolved over time, but they remain useful. In a study of 2570 characters taught in Chinese elementary schools, Shu et al. \cite{shu2003} found that 88\% of compound characters had a semantic component that was clearly related to the meaning and 62\% had a phonetic component that provided a useful cue to pronunciation. The level of phonetic regularity is greater than is often appreciated \cite{defrancis, newyorker}.

\begin{figure}
\begin{center}\includegraphics[width=0.7\textwidth]{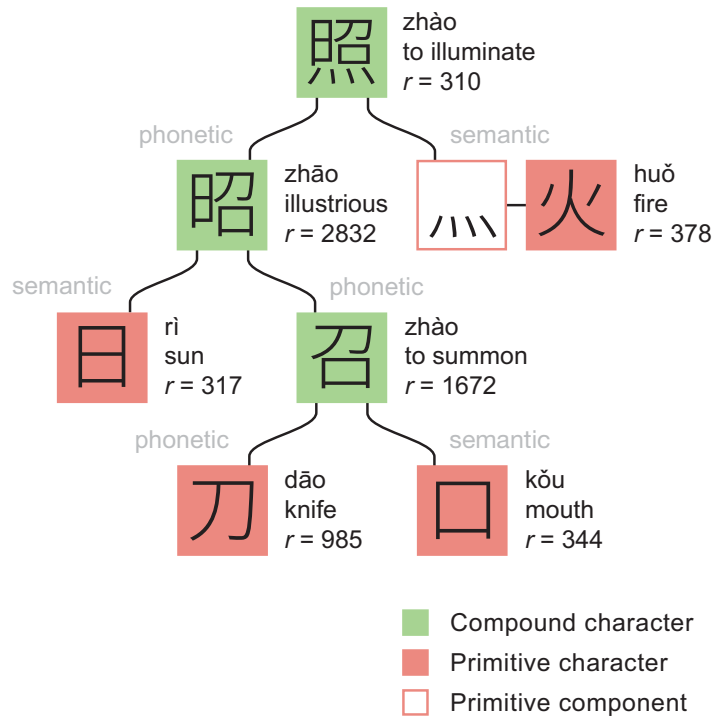}\end{center}
\caption{{\bf Structural decomposition of the character 照.} Primitive characters appear as characters in their own right whereas primitive components do not. The primitive component 灬~is an abbreviated form of the primitive character 火. The parameter $r$ is the SUBTLEX-CH usage frequency rank of the character. Pronunciations are given in pinyin romanization. Note that each character is only assigned a single meaning even though most actually possess a range of broadly related meanings.}
\label{Fig1}
\end{figure}

Compound characters are frequently used as phonetic or semantic components in other characters and so, collectively, Chinese characters form a hierarchal network. At the foundation of this network are primitive symbols, which typically originate from pictographs. Some of these primitives form characters in their own right, while others are used only as components. The structure of the character 照~(zhào, to illuminate) is shown in Fig.~\ref{Fig1}. The decomposition illustrates the typical ways in which phonetic relationships are distorted and how semantic relationships are sometimes rather general or oblique.

The semantic-phonetic structure of most Chinese characters makes the learning process somewhat different for native Chinese speakers and second language learners. When Chinese children learn to read and write they already know the spoken language and so the phonetic information can be very useful for making connections between written and spoken forms \cite{marton5}. For second language learners, who are often learning characters at a time when they know little of the language, this information is more difficult to use and the learning process is correspondingly more difficult.

But just as learning characters can be more challenging for second language learners it can also be particularly useful. The Chinese language abounds in homophones, syllables that have identical pronunciations but different meanings (it has many fewer distinct syllables compared to English, around 6 times fewer if one accounts for tones and around 20 times fewer if one neglects them). This gives a potential for ambiguity in the spoken language and acts to obscure some of the logic behind word formation. However, neither of these issues translate into written Chinese because homophones are often represented by different characters. For example, the character 照~of Fig.~\ref{Fig1} is pronounced identically to the unrelated characters 兆~(sign or portent), 罩~(cover) and 棹~(oar). Knowing characters can thus help the learner distinguish between homophones and assign distinct mental identities to the different meanings. This, in turn, can help with understanding and remembering words. For example, the verbs 照应~(zhàoyìng, to coordinate) and 照映~(zhàoyìng, to shine) are pronounced identically, but have differences in meaning that are suggested by the final characters 应~(respond) and 映~(reflect).

There is substantial debate in the literature on how characters should be taught and on the level of knowledge that is required at different educational stages \cite{allen, lam, wang1992, richardson}. This debate, as well as the importance of the problem, is reflected in the wide variety of learning methodologies found in different courses, books and apps. Here we are largely agnostic regarding the best overall approach. Rather, we consider a general question that is relevant to most of them and suggest an answer that is based on broad educational principles. The question we address is the optimum order in which Chinese characters should be learned.

There are two orders that make intuitive sense: in order of usage frequency, from high to low, and in order of network hierarchy, starting with primitives and building up compound characters using components that have already been learned. The first of these follows directly from the goal of the learner but the second merits further discussion. In general terms, the desirability of learning characters in hierarchal order follows from a broad principle of human cognition, that mastery of a complex system rests on mastery of the relevant features of its sub-components \cite{orderbook, chess}. This applies to Chinese characters if one assumes that it is productive to treat them as a complex system rather than as a set of unrelated symbols to be learned by rote. A number of experimental studies indicate that this assumption is valid. They show that orthographic awareness is of critical importance to skilled native readers and in learning to read by both Chinese children and second language learners \cite{shen2004, shen2005, marton4, wang2003, feldman1999, shu1997, connie1999}. These also show that orthographic awareness is present whether or not it is taught explicitly and, among learners, that the extent of the awareness is correlated with performance \cite{shen2007}. We consider learning characters in order of hierarchy to be desirable because we infer that a learning order that explicitly reflects orthographic principles is more likely to generate accurate and productive orthographic awareness in students.

\begin{figure}
\begin{center}\includegraphics[width=0.7\textwidth]{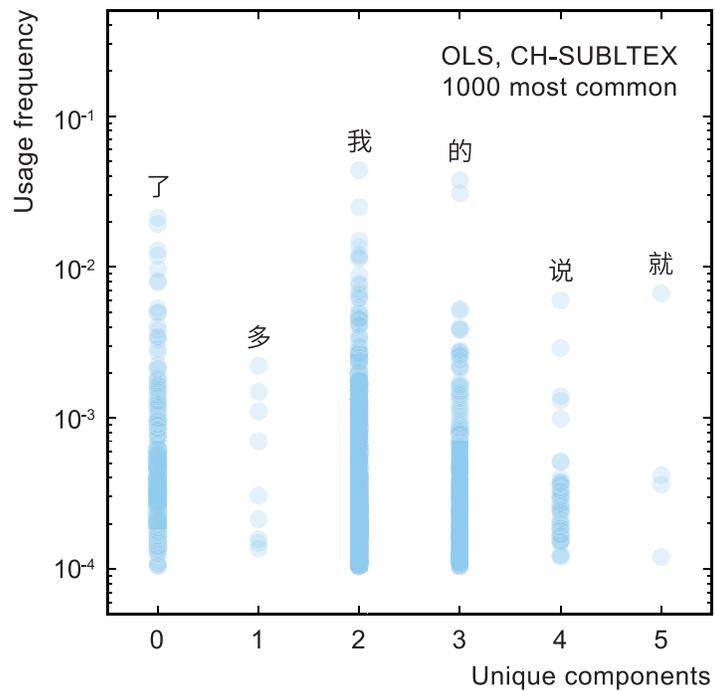}\end{center}
\caption{{\bf Usage frequency versus number of unique components for the 1000 most common Chinese characters.} This plot shows the weak relationship between character usage frequency and complexity, the latter represented by the number of unique components used to construct the character. Usage frequency is normalized to 1.0 over the whole usage frequency data set, which encompasses more characters than shown in this plot. The six characters illustrated are the most common in each column. Note that the number of unique components is not the same as visual complexity: the characters 我~and 说~have similar visual complexity (they are composed of similar numbers of strokes) but 我~is conceptually more simple, being, in the OLS character decomposition, composed of two relatively complex primitive components 手~and 戈, compared with the four from which 说~is composed.}
\label{Fig2}
\end{figure}

There is, however, necessarily a tension between learning by usage frequency and learning by hierarchy, because frequency is only weakly correlated with character complexity. This behavior can be seen in Fig.~\ref{Fig1}, where, for example, 照~appears around five times more often than its component 刀, and also in Fig.~\ref{Fig2}. Learning characters in order of frequency would therefore often mean learning characters before their components had been learned, whereas learning them in order of hierarchy would often mean learning rarer characters in advance of more common ones. 

When devising a learning order one can choose either of the extremes, of frequency or hierarchy, or attempt to find a balance between them in which some common characters are learned in advance of their components. One previous approach that searched for such a balance was a network theory-based approach by Yan et al. \cite{yan}. They demonstrated that an algorithmically-optimized, balanced order can be substantially more efficient than one that follows frequency.

Yan et al. also showed an improvement over pure hierarchal ordering, though somewhat less convincingly. It is unconvincing because they compared their optimized order to only one of many possible hierarchal orders, and there is no reason to believe that the one they choose is representative. Indeed, it will be one of the conclusions of this work that extremely efficient hierarchal orders do exist, ones that can outperform the orders produced by their algorithm.

The tension between frequency and hierarchy is a dominant consideration in determining the learning order but it is not the only one. Small-scale character-to-character patterns are also known to be important, especially for encouraging orthographic awareness \cite{ho2013, xu2014, taftchung}. Patterns can be chosen to emphasize the logic of character construction, by introducing components directly before their compounds, or to emphasize the functional role of components, by presenting their compounds in sets. These patterns are often found in human-curated orders, and especially in books on learning Chinese characters (for example, those of Heisig and Richardson \cite{heisig1, heisig2}). They embody sound educational principles, which can be understood in terms of Marton's variation theory \cite{variations1, variations2, variations3, ho2014}.

Patterns such as these are not present in orders produced by the Yan et al. algorithm. Their procedure generates a degree of character-to-character noise that means that components are rarely adjacent to the compounds that motivate their introduction and sometimes even follow them. This contrasts with the algorithm presented here, which produces orders with a high degree of logical transparency and strong clustering of related characters.

Our algorithm is built on the fundamental assumption that hierarchal orders are the pedagogically desirable way to accumulate usage frequency and we search among this subset of orders for the one that is most efficient. The algorithm is implemented using the conceptual framework of network theory, within which we conceive the network of Chinese characters as a directed analytic graph \cite{li}. The nodes in the graph represent characters and the edges represent the structural relationships between them. We devise a measure of node centrality that relates each character's usage frequency to the effort required to learn it, and order the characters by this measure to provide a first approximation to the optimal learning order. We then sort this list into topological (hierarchal) order using an algorithm designed to minimally disturb the starting order. The algorithm can be applied to any scheduling task where nodes have intrinsic differences in importance and must be visited in topological order.

Following this introduction, we describe the structure of our algorithm in detail, including how we define learning efficiency and how we calculate the cost of learning characters. We discuss the robustness of the algorithm and study the characteristics of the orders it produces.

In the final section we apply our algorithm to a network that is expanded to include Chinese words. Chinese words can be single characters but they are more frequently compounds of two or more. They are the primary units of communication in the Chinese language and so characters, rather like letters in alphabetic scripts, may be considered useful only in so far as they help to build words or act as words themselves. Reflecting this, words can be moved to center stage and, instead of having character usage frequencies drive the acquisition of components, word usage frequencies can be used to drive the acquisition of characters and their components. We explore the results of this more holistic approach.


\section*{Analysis}

\subsection*{Overview.}

The network of Chinese characters can be represented as a directed analytic graph. Nodes represent characters, with their visual forms, pronunciations and meanings, and edges represent the structural relationships between characters and the nature of those relationships, whether semantic, phonetic or otherwise. Learning Chinese characters means memorizing a productive subset of this network. 

Our aim is to derive a character learning order which maximizes learning efficiency. Such an order maximizes cumulative usage frequency while minimizing the effort required to learn it. To this end, we assign a usage frequency to each character along with an estimate for the effort required to learn it, its \textit{learning cost}. Learning costs are calculated using a model that assumes that characters are learned in hierarchal order.

We incorporate usage frequency and learning cost into a measure of character centrality. This measure indicates the relative importance of the character to the learner, prioritizing usage frequency and penalizing learning cost. Ordering characters by this centrality provides a first approximation to the final learning order. This order is approximate because ordering by centrality does not imply ordering by hierarchy, which must be imposed in a separate step.

Hierarchal ordering is imposed using an algorithm designed to topologically sort our centrality-ordered list in a way that minimally disturbs it. Higher-centrality characters are learned first only when allowed topologically. 

The algorithm can easily account for characters that are already known to the learner (their learning costs can be set to zero) or characters that are partially known (their learning costs can be suppressed). This capability could be useful in software applications, which could dynamically update the learning order as the student progresses.

The algorithm has potential applications beyond the learning of Chinese characters, and can be applied to any scheduling task where nodes have intrinsic differences in importance yet must be visited in topological order.

\subsection*{Learning efficiency.}

\begin{figure}
\begin{center}\includegraphics[width=0.7\textwidth]{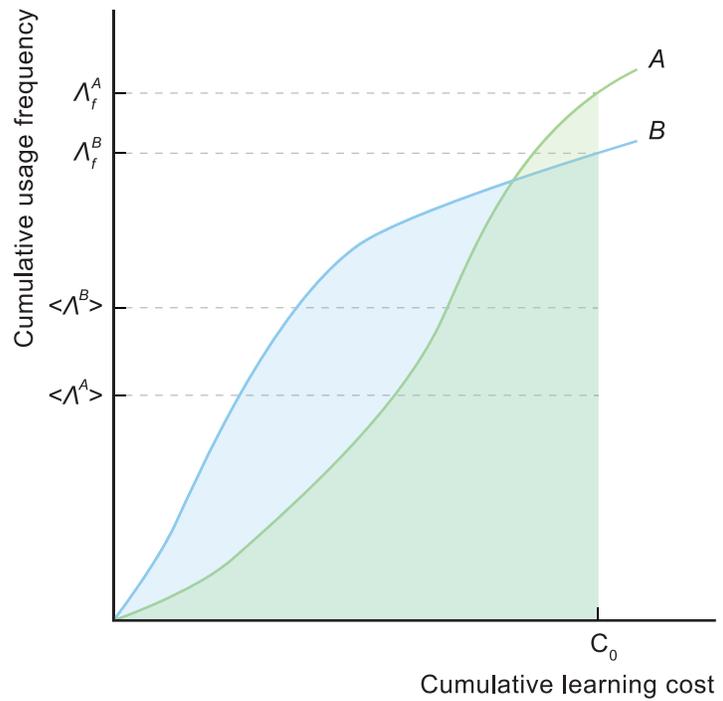}\end{center}
\caption{{\bf Measures of learning efficiency.} The curves $A$ and $B$ represent two different learning curves. For each curve, the final learning efficiency $\Lambda_f$ is the cumulative usage frequency for a specific cumulative learning cost $C_0$, and the integral learning efficiency $\langle\Lambda\rangle$ is the average cumulative usage frequency between the origin and $C_0$. Curve $A$ has higher $\Lambda_f$ but lower $\langle\Lambda\rangle$. Illustrated values for $\langle\Lambda\rangle$ are approximate.}
\label{Fig3}
\end{figure}

A typical learning scenario is characterized by a fixed available effort, with which the learner seeks to acquire the maximum cumulative usage frequency as rapidly as possible. The learning process can be visualized as a \textit{learning curve} in a space defined by axes of cumulative usage frequency and cumulative learning cost. This is illustrated in Fig.~\ref{Fig3}. Efficient learning curves rise quickly and reach high end-points.

Learning curves for different orders can be compared visually, as in the figure, but it is convenient to parameterize them. We propose a two-parameter scheme.

The first parameter is the cumulative usage frequency at the end of the learning process, once the maximum cumulative learning cost $C_0$ is reached. We call this parameter the \textit{final learning efficiency} $\Lambda_f$. High $\Lambda_f$ is characteristic of efficient learning curves. Note that when comparing curves, it is necessary to use the same usage frequencies for characters that appear in both curves, even though the curves may not cover identical sets of characters. To ensure this, we normalize the entire usage frequency data set to 1.0, which becomes the maximum possible value of $\Lambda_f$.

The second parameter concerns how the maximum cumulative usage frequency is approached. Consider the two curves shown in Fig.~\ref{Fig3}. The curve $A$ has higher $\Lambda_f$ but, over much of the learning process, actually performs less well than curve $B$. Curve $A$ might be one that prioritizes longer-term cumulative frequency at the cost of shorter-term. The difference would be immaterial if the learning process had a short extent in time but this is not typically the case. Learning may take place over many months, during which the learner would likely be exposed to other parts of the language. In this case it might make sense to have a learning curve that rises quickly, even at the cost of some longer-term cumulative usage frequency. We parameterize this using the average cumulative usage frequency, which we call the \textit{integral learning efficiency} $\langle\Lambda\rangle$ and calculate using

\begin{equation}\label{eq:efficiency}
\langle\Lambda\rangle = \frac{1}{C_0} \int_0^{C_0} F(C) \,\mathrm{d}C
\end{equation}
where $F$ is the cumulative usage frequency and $C$ is the cumulative learning cost. 

\subsection*{Centrality.}

We define the centrality $\eta_i$ of character $i$ to be

\begin{equation}\label{eq:centrality}
\eta_i = \frac{f_i}{c_i},
\end{equation}
where $f_i$ is the usage frequency and $c_i$ is the learning cost. This quantity is the ratio of the benefit and cost that each character represents to the learner. Learning characters in order of $\eta$ will therefore tend to satisfy the prime concern of the learner, of maximizing cumulative usage frequency and minimizing effort. These learning curves will rise steeply and have high end points, or, in the language of the previous section, be characterized by high integral and final learning efficiencies.

Values for ${f_i}$ can be extracted from corpora of written Chinese. Values for ${c_i}$ are more difficult to assign objectively and we estimate them using a learning model. In our model we use different procedures to assign costs to primitives and compound characters.

The cost $c^{(p)}_i$ of learning a primitive $i$ is taken to be
\begin{equation}\label{eq:alpha}
c^{(p)}_i= 1 + \gamma s_i
\end{equation}
where $s_i$ is the number of strokes that make up the character. Using a $\gamma$ of 0.1 would, for example, mean the cost of learning 口~is 1.3 and the cost of learning 豕~is 1.7. This is a simple approximation to the true learning cost, which would depend on a variety of other things, and likely in complex ways: the learner's familiarity with the strokes that make up the character, their knowledge of similar primitives, and the visual similarity between the primitive and the thing it represents. 

The cost $c^{(c)}_i$ of learning a compound $i$ is taken to be
\begin{equation}\label{eq:beta}
c^{(c)}_i= m_i
\end{equation}
where $m_i$ is number of combinations used to build the character. Thus, the cost of learning 的~would be 1, because it is a compound of two components 白~and 勺, and the cost of 茶~would be 2, because it is a compound of three (艹, 人~and 木). For characters that are variants of others (such as 灬~, which is a variant form of 火) we assign a cost of 1. We do not take special account of characters with repeating elements (such as 品) for which we likely overstate the cost. 

In this work we take $\gamma= 0.1$. This means that the cost of learning the simplest primitive would be similar to the cost of learning the simplest compound. A typical primitive would be around twice as difficult. Our final learning order depends on the specific value chosen for $\gamma$ but our conclusions are the same for any reasonable value.

Our learning cost model assumes that characters are learned in hierarchal order. When we calculate the cost of learning a compound character we do not include the cost of learning the components themselves, which we assume have already been learned. 

Our model implies that the total cost of learning a fixed set of characters is identical for all hierarchal orders. All final learning efficiencies will be identical, with the only differences being in the integral learning efficiencies.

\subsection*{Topological sort.}

Learning characters in order of centrality prioritizes characters that are useful and easy to learn but it does not ensure that characters are ordered according to the character hierarchy. For example, the simple and common 的~will appear a long way in advance of its components 白~and 勺, which appear much less frequently as characters.

We resolve this issue with a sorting algorithm that modifies the centrality-ordered list to ensure that all characters appear before those in which they act as components. The algorithm is illustrated in Fig.~\ref{Fig4} and may be described as follows:

\begin{figure}
\begin{center}\includegraphics[width=0.7\textwidth]{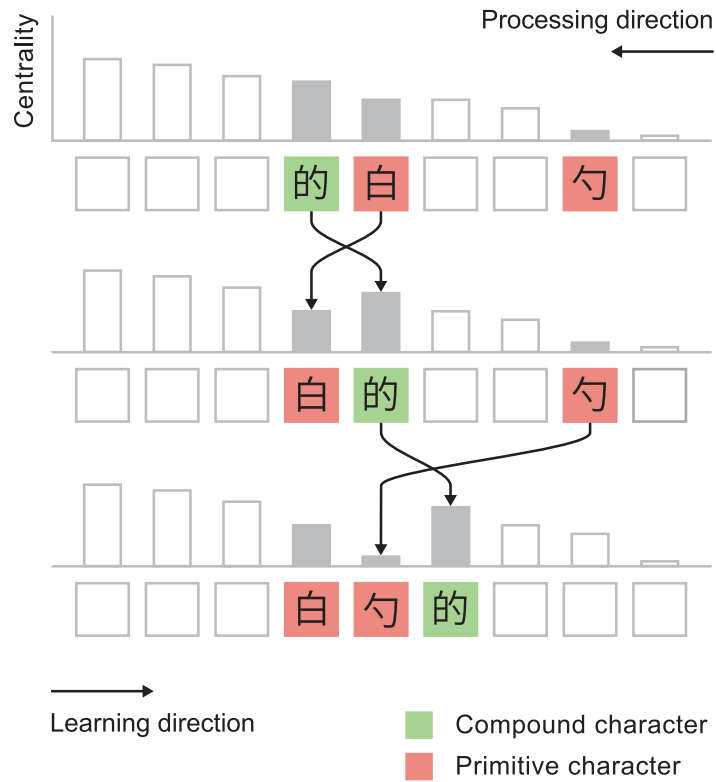}\end{center}
\caption{{\bf Illustration of the topological sort algorithm.} The ordered list is processed from low to high centrality (right to left in the figure). Once 的~is reached, its components are checked in turn. 白~is found to lie to the right of 的~and so is repositioned to its left. Likewise 勺~is found to the right of 的~and is similarly repositioned. 勺~is positioned to the right of 白~because it has lower centrality. The centralities used in this figure are for illustrative purposes only.}
\label{Fig4}
\end{figure}

\begin{enumerate}
\item Process characters from low to high centrality (right to left in the figure). This is opposite to the order in which characters are learned.
\item Decompose each character into a list of its primitives and all intermediate characters. For example, 照~should be decomposed into 昭, 召, 日, 刀, 口~and 灬.
\item Determine the position of each component in the centrality-ordered list. If the position is to the left of the character then no action is taken. If it is to the right of the character then it should be moved the character's left. Move the character as far left as it will go, while still remaining to the right of all characters with higher centrality.
\end{enumerate}

\begin{figure}
\begin{center}\includegraphics[width=0.7\textwidth]{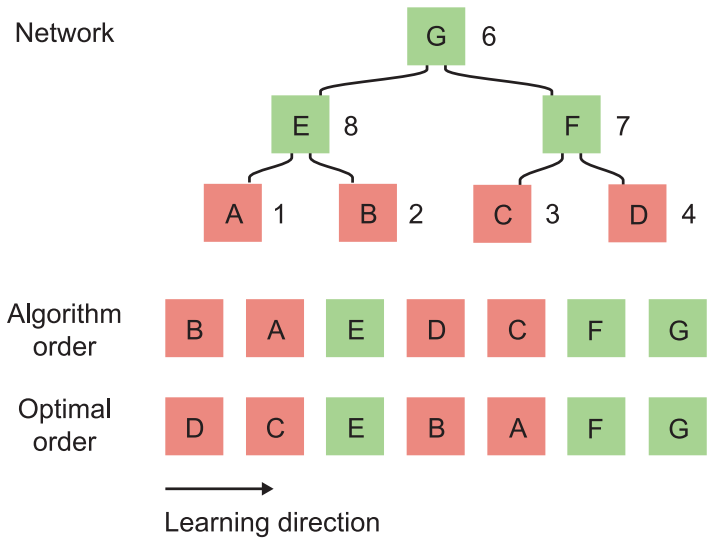}\end{center}
\caption{{\bf A network where our algorithm does not return the optimal character order.}  A hypothetical network where the integral learning efficiency of the order generated by the algorithm is lower than another possible order. Letters represent Chinese characters (for example, E is a compound character formed from primitives A and B) and the numbers are centralities. Both orders have identical final learning efficiencies.}
\label{Fig5}
\end{figure}

This procedure ensures that characters are relocated only when necessary and always to the optimum position within the region allowed by the hierarchy. This results in a highly optimized order. However, it is not necessarily the most efficient order possible and we can find special cases, such as the network in Fig.~\ref{Fig5}, where the algorithm does generate a sub-optimal order. We have not found such instances in the real character network but cannot prove that they do not exist.

\subsection*{Source data.}

We use two different representations of the simplified Chinese character network, one compiled with an emphasis on etymological correctness and one with an emphasis on the visual relationships between characters. The former is based on a preliminary version of forthcoming dictionary by \textit{Outlier Linguistic Solutions} \cite{outlier} and the latter is taken from the books \textit{Remembering Simplified Hanzi 1 and 2} by Heisig and Richardson. We refer to these networks as the OLS and HR networks, respectively. The networks have similar coverage: OLS covers 3507 characters and primitives, and HR covers 3250, with 2990 in common between them. The majority of the decompositions are identical and the majority of the differences originate from decisions regarding encoding (a number of components do not have Unicode code points and others can reasonably be represented by more than one code point).

Usage frequency data for characters and words are taken from the SUBTLEX-CH database \cite{subtlex}, which is derived from Chinese film and television subtitles. We chose this database because it is comprehensive and is representative of modern colloquial Chinese. In any practical application of our algorithm, frequency data should be chosen with the specific goals of the learner in mind. The database contains 5938 unique characters and 99121 unique words with frequencies calculated from a total corpus of 46.8 million characters (algorithmically segmented into 33.5 million words). All usage frequencies used in this study are normalized to the whole database. Normalizing in this way, both the OLS and HR networks have cumulative usage frequencies of 0.992. 

Stroke counts were taken from the UNIHAN database using the python package cjklib \cite{cjklib}. Where stroke data was unavailable, the number was set to zero.

\section*{Results and discussion}

\begin{figure}
\begin{center}\includegraphics[width=1.0\textwidth]{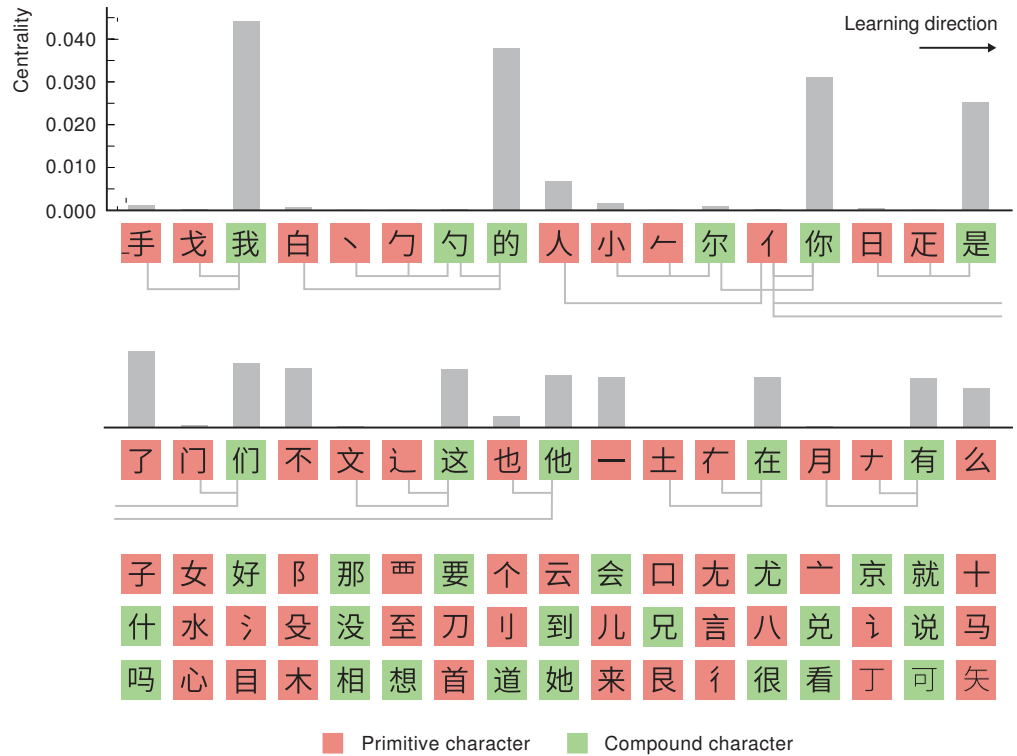}\end{center}
\caption{{\bf The first 85 characters of our optimized learning order.} Taken together these characters have a cumulative usage frequency of 0.42.}
\label{Fig6}
\end{figure}

\begin{figure}
\begin{center}\includegraphics[width=0.7\textwidth]{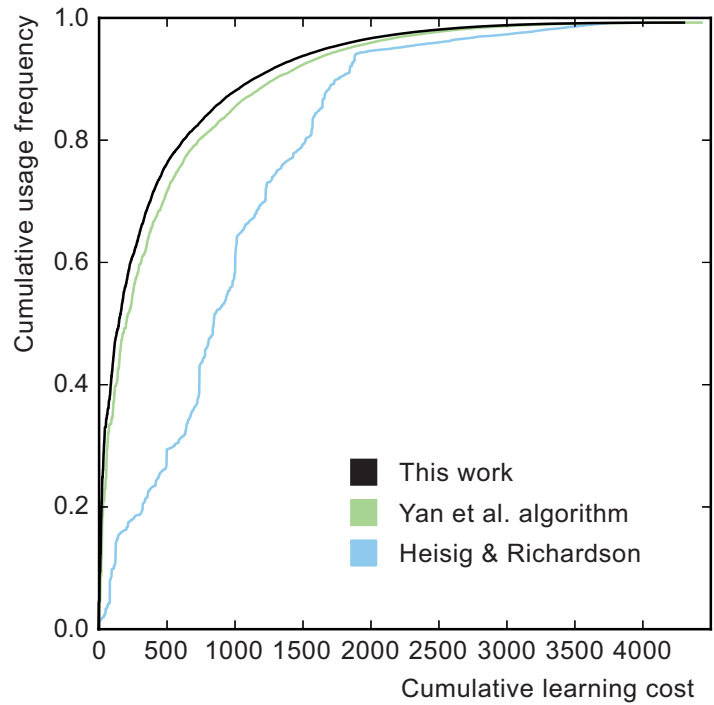}\end{center}
\caption{{\bf Learning curves.} The black and green curves were created using the OLS character decompositions and the two different learning order algorithms. The Yan et al. algorithm was optimized up to a cumulative learning cost of $C_0 = 4000$. The blue curve uses the HR network with Heisig and Richardson's fixed character order. Learning efficiencies are presented in Table~\ref{Table}.}
\label{Fig7}
\end{figure}

\begin{figure}
\begin{center}\includegraphics[width=1.0\textwidth]{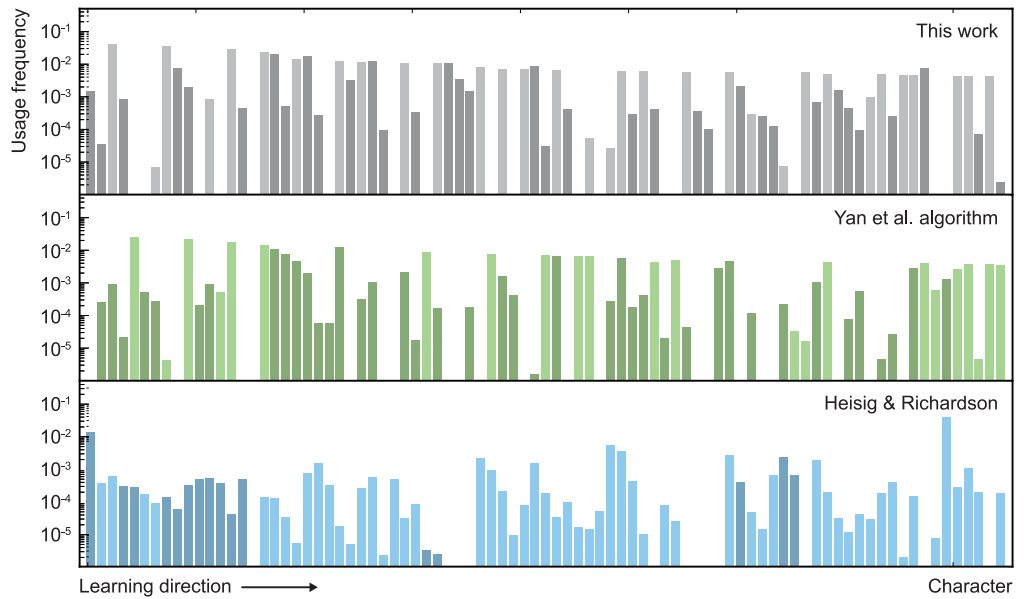}\end{center}
\caption{{\bf Usage frequencies for the first 85 characters.} The gray, green and blue bars correspond to the black, green and blue curves in Fig.~\ref{Fig8}. Dark bars represent primitives and light bars represent compounds.}
\label{Fig8}
\end{figure}

Fig.~\ref{Fig6} shows the first 85 characters from the learning order derived using the OLS network. The full learning curve for the OLS network is shown in Fig.~\ref{Fig7}, where it is compared to the Yan et al. algorithm and to the fixed character order of Heisig and Richardson. Fig.~\ref{Fig8} shows usage frequencies for the first 85 characters of each of the curves in Fig.~\ref{Fig7}. Learning efficiencies are presented in Table~\ref{Table}.

The shape of the Heisig and Richardson curve in Fig.~\ref{Fig7} can be understood from the structure of their book. The first half of the curve, between the origin and the large discontinuity, covers their first volume, in which they introduce the bulk of the primitives. These are grouped in chapters according to meaning and each one is followed by all the high-usage compound characters that can be made at that point. This explains the alternating pattern of sharp upward jumps and gentle slopes. The second volume introduces the lower-frequency compounds that are not included in the first. The authors present a hierarchal order which aims for a relatively high $\Lambda_f$ by the end of the first volume but with no particular regard for the shape of the curve. Note that their curve in Fig.~\ref{Fig7} was calculated using SUBTLEX-CH frequency data, which may differ from the frequency data which they used to select their characters and order them.

The curves corresponding to our algorithm and that of Yan et al. were calculated using identical character networks and usage frequencies. We also used identical learning models in order to make a properly normalized comparison. To account for the Yan et al. order being non-hierarchal we extended our learning model to include the cost of any unlearned components, making it similar to the model used in their publication. 

We find that our algorithm gives better $\Lambda_f$ and $\langle\Lambda\rangle$. This follows largely from the fact that our order is hierarchal and so there is no inefficiency associated with learning characters by rote and then later re-learning some of the components. But it is also dependent on the particular characteristics of the Chinese character network, because there is no a priori reason why the optimal learning order need be hierarchal. For example, in the extreme, hypothetical case that a small number of complex characters accounted for the vast majority of usage frequency it would be more efficient to ignore the components and just learn them by rote. With such a network the Yan et al. algorithm might perform better because it has access to the non-hierarchal parts of the character order space. Indeed, it remains possible that that a non-hierarchal order is optimal for the character network as it exists.

Our algorithm produces orders that exhibit a high degree of logical transparency. This behavior can be seen in Fig.~\ref{Fig6}, and it follows directly from the algorithm, which tends to cluster components directly before the compounds in which they are used. The sequence 心, 目, 木, 相, 想~is a typical example. Similar sequences appear frequently in the Heisig and Richardson order but rarely in the orders produced by the Yan et al. algorithm.

The Heisig and Richardson order is characterized by the introduction of compound characters in sets that have a particular primitive in common. The pedagogical advantages of this pattern were discussed in the introduction but are not fully realized because of the absence of phonetic information (they do not give character pronunciations) and their deliberate strategy of assigning semantic values to all components, whether or not this is etymologically correct. 

\begin{figure}
\begin{center}\includegraphics[width=0.7\textwidth]{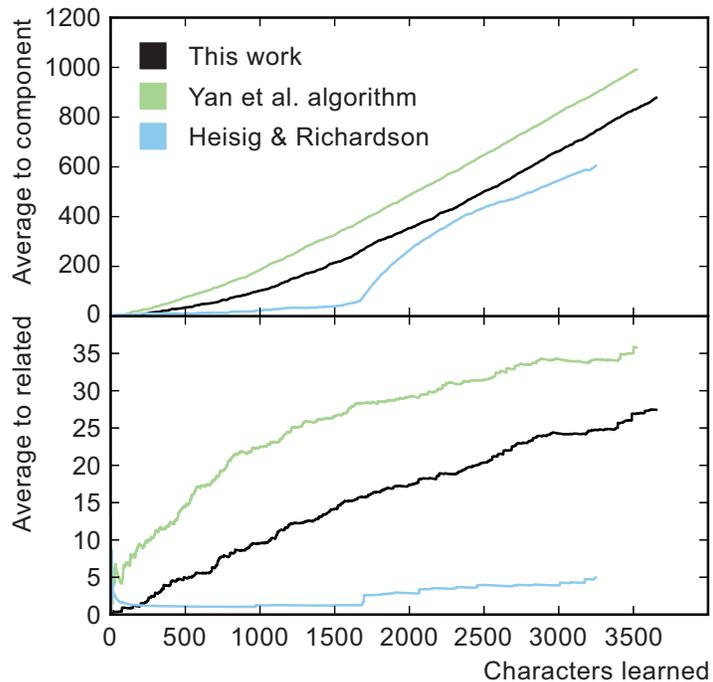}\end{center}
\caption{{\bf Measures of character clustering.} The top panel shows the average distance, in number of characters, to the closest preceding component. The bottom panel shows the average distance, in number of characters, to another character which shares a component. Curves were generated with a fixed cumulative learning cost of $C_0 = 4000$. Averages below 250 characters are not shown because in this region the averages fluctuate wildly.}
\label{Fig9}
\end{figure}

Grouping of characters into meaningful sets does not occur in either of the algorithmically-generated orders and its absence offers an avenue for improvement. Nevertheless, the two algorithms do produce orders with markedly different degrees of clustering between related characters (of which grouping of characters into sets might be considered a limiting case). Fig.~\ref{Fig9} shows all three orders compared using two parameters which measure the degree of clustering: the average distance, in number of characters, of each character to its closest preceding component and to the closest character sharing a component. In both these measures the Heisig and Richardson order exhibits the strongest clustering, with ours intermediate between theirs and Yan et al. Along with the logical transparency of our order, we take this to indicate improved pedagogical characteristics compared to Yan et al.

In summary, we have shown that our algorithm can identify pure hierarchal orders of Chinese characters that give high learning efficiencies. The numerical improvements over Yan et al. are modest, but they are coupled with character-to-character patterns that we suggest are pedagogically advantageous: the order is strictly hierarchal, components are typically introduced immediately before they are used, and there is stronger clustering of related characters. 

\begin{table}
\caption{
{\bf Learning curve parameters.} The number of characters learned $N$, final learning efficiency $\Lambda_f$, and integral learning efficiency $\langle\Lambda\rangle$ for reference cumulative learning costs of $C_0 = 500$ and $C_0 = 1500$. The Yan et al. algorithm was optimized up to a cumulative learning cost of $C_0 = 4000$.}
\begin{tabular}{| l | l | l | c | c | c | c | c | c |}
\hline
{\bf Curve}  & {\bf Network} & {\bf Algorithm} & \multicolumn{3}{|c|}{ {\bf $C_0 = 500$} } & \multicolumn{3}{|c|}{ {\bf $C_0 = 1500$} } \\ \hline
  & &  & {\bf $N$}  & {\bf $\Lambda_f$}  & {\bf $\langle\Lambda\rangle$} & {\bf $N$}  & {\bf $\Lambda_f$}  & {\bf $\langle\Lambda\rangle$} \\ \hline
This work    & OLS  &  This work   &  432  &   0.760   &  0.560  &  1333  &  0.938  &  0.770  \\ 
Yan et al.    & OLS  &  Yan et al.    &  374  &   0.714   &  0.501  &  1199  &  0.923  &  0.740  \\  
HR              & HR    &  N/A             &  444  &   0.267   &  0.163  &  1335  &  0.793  &  0.432  \\ \hline
\end{tabular}
\begin{flushleft} 
\end{flushleft}
\label{Table}
\end{table}

\subsection*{Words}

We can expand our analysis to include multiple-character Chinese words by making minor changes to the network and learning model. The network is expanded using multiple-character words from the SUBTLEX-CH word database (limited to the 10000 most common words, for computational convenience). This database is also used for all usage frequencies, including those of characters. Previously, when using character frequencies, the 知~and 道~of 知道~(zh\={\i}dào, to know) had very high frequencies because the compound is common, but when we switch to word frequencies their frequencies become very small compared to the compound - 知道~becomes the important thing to learn and 知~and 道~are only learned in advance because they support it in the hierarchy. The learning model is expanded to account for multiple-character words by assigning them learning costs equal to the number of character combinations required to build them. Thus 知, 道~and 知道~all have learning costs of 1.

The learning order calculated in this way is the learning order for the productive units of communication. Characters and their components are introduced in hierarchal order only when needed to build multiple-character words or when the character forms a single-character word itself. This approach to learning to read and write Chinese has the advantage that the things being learned - the words - can be put to immediate and productive use in reading and writing sentences, activities that helps the learning process. This is not the case for characters, which often only acquire their usage frequencies via expression through words. 知道~is a categorically useful word to know, that can be assigned a clear mental identity and used immediately. Its component characters 知~and 道~are rarely used alone as words and take their character usage frequencies primarily from their presence in this and other, less common compounds. In the character learning order they appear as unrelated characters, ten characters apart, yet 道~can't be used by a learner until 知~has already been learned. In the word learning order they appear together in the logical sequence 首, 道, 矢, 知, 知道.

\begin{figure}
\begin{center}\includegraphics[width=0.7\textwidth]{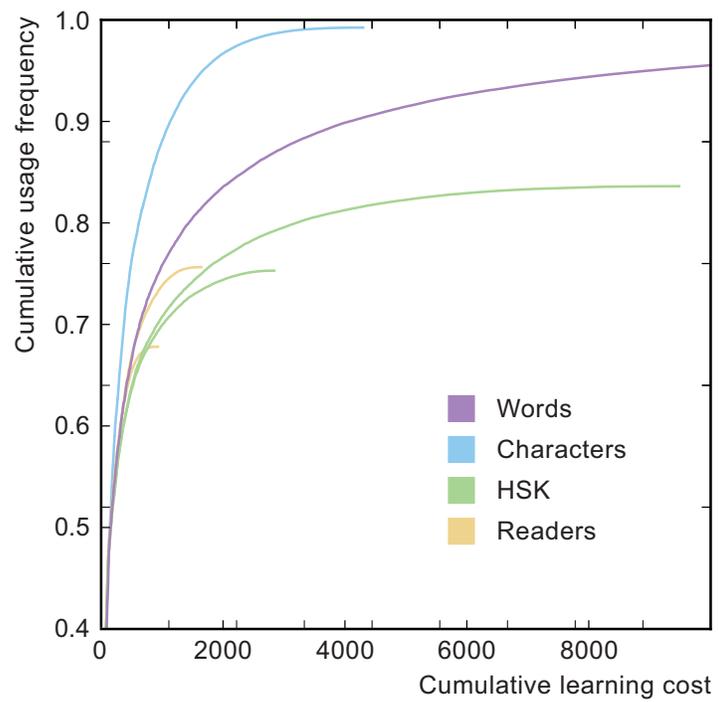}\end{center}
\caption{{\bf Learning curves for characters and words.} The green curves correspond to HSK word lists for levels 1 to 4 (shorter curve) and 1 to 6 (longer curve). The yellow curves correspond to word lists generated from two levels of beginner readers. All curves were created using the OLS character decompositions.}
\label{Fig10}
\end{figure}

The word learning curve is shown in Fig. \ref{Fig10}, where it is compared to the curve for characters. The difference between the two is stark, indicating that mastery of characters is substantially easier than mastery of the words they combine to create (something which accords with learner experience).

The other curves in the figure show what happens to the word curve when the target set of words is a subset of the wider language. These curves represent realistic situations for the application of our algorithm, in which a student is trying to master the vocabulary required to pass a course or read a particular book. The figure shows curves for the vocabulary lists for levels 1-4 and 1-6 of the HSK Chinese Proficiency Test. This is an exam administered by the Chinese National Office for Teaching Chinese as a Foreign Language (NOTCFL) \cite{hsk} and the lists contain 1200 and 5000 words, respectively. We also show curves corresponding to the lowest two levels of a series of Chinese readers, containing 496 and 977 distinct words, respectively \cite{mc1, mc2}. The text from the readers was segmented into words using the same algorithm used to calculate the word usage frequency list (implemented using PyNLPIR \cite{segment}). These curves necessarily have inferior $\Lambda_f$ and $\langle\Lambda\rangle$ compared with the curve for the wider language but nevertheless represent efficient approaches to the restricted goals. 

The algorithm would be similarly useful in managing the transition between different levels of a course or from one book to another. In these situations, the algorithm would provide an efficient way to bridge the gap in vocabulary.

\section*{Supporting Information}


\paragraph*{S1 File.}
\label{S1_File}
{\bf Input files and character orders.} data.zip is a zipped file containing input data (usage frequencies, decompositions, stroke numbers, target word lists) and final character and word orders. A README within the file contains information on the origin of the input data and guidance on use of the output orders.

\section*{Acknowledgments}

This work was supported by the Shanghai Key Lab for Particle Physics and Cosmology (SKLPPC), Grant No. 15DZ2272100. We acknowledge the generosity of Outlier Linguistic Solutions in sharing part of their dictionary prior to publication.

\nolinenumbers

%
%
%

\end{CJK*}


\begin{thebibliography}{10}

\bibitem{ke2001} Ke C, Wen X, Kotenbeutel C. Report on the 2000 CTLA Articulation Project. Journal of the Chinese Language Teachers Association. 2001; 36(3): 25-60. 

\bibitem{walker1989} Walker GLR. Intensive Chinese Curriculum: The EASLI Model. Journal of the Chinese Language Teachers Association. 1989; 24(2): 43-83. 

\bibitem{light1975} Light T. Controlled Composition and Reading. Journal of the Chinese Language Teachers Association. 1975; 10(2): 70-79. 

\bibitem{kane} Kane D. The Chinese Language: Its History and Current Usage. Tuttle Publishing; 2006. 

\bibitem{marton1} Tse SK, Cheung WM. Chinese and the Learning of Chinese. In: Marton F, Tse SK, Cheung WM, editors. On the Learning of Chinese. Sense Publishers; 2010. pp. 1-8.

\bibitem{shu2003}  Shu H, Chen X, Anderson RC, Wu N, Xuan Y. Properties of School Chinese: Implications for learning to read. Child Development. 2003; 74(1): 27-47.

\bibitem{defrancis} DeFrancis J. Visible Speech: The Diverse Oneness of Writing Systems. University of Hawaii Press; 1989. 

\bibitem{newyorker} Chiang T. Bad Character. The New Yorker. 16 May 2016.

\bibitem{marton5} Tse SK, Marton F, Ki WW, Loh EKY. Learning Characters. In: Marton F, Tse SK, Cheung WM, editors. On the Learning of Chinese. Sense Publishers; 2010. pp. 75-102.




\bibitem{wang1992} Wang AY, Thomas MH. The Effect of Imagery-Based Mnemonics on the Long-Term Retention of Chinese Characters. Language Learning. 1992; 42(3): 359-376.

\bibitem{richardson} Richardson TW. Chinese Character Memorization and Literacy: Theoretical and Empirical Perspectives on a Sophisticated Version of an Old Strategy. In: Guder A, Xin J, Yexin W, editors. The Cognition, Learning and Teaching of Chinese Characters. Beijing Language University Press; 2007.

\bibitem{allen} Allen JR. Why Learning to Write Chinese is a Waste of Time: A Modest Proposal. Foreign Language Annals. 2008; 41(2): 237-251.

\bibitem{lam} Lam HC. A Critical Analysis of the Various Ways of Teaching Chinese Characters. Electronic Journal of Foreign Language Teaching. 2011; 8(1): 57-70.




\bibitem{orderbook} Ritter FE, Nerb J, Lehtinen E, O'Shea T, editors. In Order to Learn: How the Sequence of Topics Influences Learning. Oxford University Press; 2007. 

\bibitem{chess} Chase WG, Simon HA. Perception in Chess. Cognitive Psychology. 1973; 4: 55-81.







\bibitem{shen2004} Shen HH. Level of Cognitive Processing: Effects on Character Learning Among Non-Native Learners of Chinese as a Foreign Language. Language and Education. 2004; 18(2): 167-182.

\bibitem{shen2005} Shen HH. An Investigation of Chinese-Character Learning Strategies Among Non-Native Speakers of Chinese. System. 2005; 33: 49-68.

\bibitem{wang2003} Wang M, Perfetti CA, Liu Y. Alphabetic Readers Quickly Acquire Orthographic Structure in Learning to Read Chinese. Scientific Studies of Learning. 2003; 7(2): 183-208.

\bibitem{feldman1999} Feldman LB, Siok WWT. Semantic Radicals Contribute to the Visual Identification of Chinese Characters. Journal of Memory and Language. 1999; 40: 559-576.

\bibitem{shu1997} Hua S, Anderson RC. Role of Radical Awareness in the Character and Word Acquisition of Chinese Children. Reading Research Quarterly. 1997; 32(1): 78-89.

\bibitem{marton4} Lam HC. Orthographic Awareness. In: Marton F, Tse SK, Cheung WM, editors. On the Learning of Chinese. Sense Publishers; 2010. pp. 53-73.

\bibitem{connie1999} Ho CSH, Wong WL, Chan WS. The Use of Orthographic Analogies in Learning to Read Chinese. Journal of Child Psychology and Psychiatry. 1999; 40(3): 393-403.






\bibitem{shen2007} Shen HH, Ke C. Radical Awareness and Word Acquisition Among Nonnative Learners of Chinese. The Modern Language Journal. 2007; 91(1): 97-111.


\bibitem{yan}
Yan X, Fan Y, Di Z, Havlin S, Wu J. Efficient Learning Strategy of Chinese Characters Based on Network Approach. PLoS ONE. 2013; 8: e69745. doi:10.1371/journal.pone.0069745




\bibitem{xu2014} Xu Y, Perfetti CA, Chang LY. The Effect of Radical-Based Grouping in Character Learning in Chinese as a Foreign Language. The Modern Language Journal. 2014; 98(3): 773-793.

\bibitem{taftchung} Taft M, Chung K. Using Radicals in Teaching Chinese Characters to Second Language Learners. Psychologia. 1999; 42: 243-251.

\bibitem{ho2013} Ho CL, Bik MAT. Drawing on the Variation Theory to Enhance Students' Learning of Chinese Characters. Instructional Science. 2013; 41: 955-974.



\bibitem{heisig1}
Heisig JW, Richardson TW. Remembering Simplified Hanzi: Book 1, How Not to Forget the Meaning and Writing of Chinese Characters. University of Hawaii Press; 2008.

\bibitem{heisig2}
Heisig JW, Richardson TW. Remembering Simplified Hanzi: Book 2, How Not to Forget the Meaning and Writing of Chinese Characters. University of Hawaii Press; 2012.



\bibitem{variations1} Marton F, Booth S. Learning and Awareness. Lawrence Eribaum Associates; 1997.

\bibitem{variations2} Marton F, Pang MF. On Some Necessary Conditions of Learning. The Journal of the
Learning Sciences. 2006; 15(2): 193-220.

\bibitem{variations3} Ling LM, Marton F. Towards a Science of the Art of Teaching: Using Variation Theory as a Guiding Principle of Pedagogical Design. International Journal for Lesson and Learning Studies. 2012; 1(1): 7-22.

\bibitem{ho2014} Ho CL. Elaborating the Concepts of Part and Whole in Variation Theory: The Case of Learning Chinese Characters. Scandinavian Journal of Educational Research. 2014; 58(3): 337-360.






\bibitem{li}
Li J, Zhou J. Chinese Character Structure Analysis Based on Complex Networks. Physica A. 2007; 380. 



\bibitem{outlier} Henson A, Renfroe J. Outlier Dictionary. Outlier Linguistics Solutions USA, LLC; 2015.






\bibitem{subtlex} Cai Q, Brysbaert M. SUBTLEX-CH: Chinese Word and Character Frequencies Based on Film Subtitles. PLoS ONE. 2010; 5: e10729. doi: 10.1371/journal.pone.0010729





\bibitem{cjklib} cjklib 0.3.2 Available: https://pypi.python.org/pypi/cjklib. Accessed: 1 May 2016.


\bibitem{hsk} HSK Chinese Proficiency Test. Available: http://www.chinesetest.cn. Accessed: 16 January 2016.

\bibitem{mc1} Burnett FH. The Secret Garden: Mandarin Companion Graded Readers Level 1. Mind Spark Press LLC; 2015.

\bibitem{mc2} Dickens C. Great Expectations (Parts 1 and 2): Mandarin Companion Graded Readers Level 2. Mind Spark Press LLC; 2016.


\bibitem{segment} PyNLPIR 0.4.4 Available: https://pypi.python.org/pypi/PyNLPIR. Accessed: 1 May 2016.


\end{thebibliography}
\end{document}